\documentclass[11pt,a4paper]{article}

\usepackage{AACL-IJCNLP2020}

\usepackage{times}
\usepackage{url}
\usepackage{latexsym}

\usepackage{microtype}

\aclfinalcopy 

\newcommand{\statsys}{\texttt{NLPStatTest}} 

\usepackage{amsmath}
\usepackage{graphicx}
\usepackage{amssymb}
\usepackage{amsthm}

\theoremstyle{definition}
\newtheorem{definition}{Definition}

\usepackage{algorithm}

\usepackage{bm}

\title{\statsys: A Toolkit for Comparing NLP System Performance}

\author{Haotian Zhu \hspace{0.2in} Denise Mak \hspace{0.2in} Jesse Gioannini
  \hspace{0.2in} Fei Xia\\
  University of Washington, Seattle, USA\\
  {\tt \{haz060, dpm3, jessegio, fxia\}@uw.edu}}

\date{}

\begin{document}
\maketitle
\begin{abstract}
  Statistical significance testing centered on $p$-values is commonly used to compare NLP system performance, but $p$-values alone are insufficient because statistical significance differs from practical significance. The latter can be measured by estimating effect size. 
  In this paper, we propose a three-stage procedure for comparing NLP system performance and provide a toolkit, \statsys, that automates the process.
  Users can upload NLP system evaluation scores and the toolkit will analyze these scores, run appropriate significance tests, estimate effect size, and conduct power analysis to estimate Type II error.
  The toolkit provides a convenient and systematic way
  to compare NLP system performance that goes beyond statistical significance
  testing.
  
\end{abstract}

\section{Introduction}
\label{intro}

In the field of natural language processing (NLP), the common practice is to use statistical significance testing \footnote{Here we adopt the frequentist approach to hypothesis testing. The debate over frequentist and Bayesian is beyond the scope of this paper.} to demonstrate that the improvement exhibited by a proposed system over the baseline reflects meaningful differences
, not happenstance \cite{ref3,ref60}. The American Statistical Association emphasizes that 
{\emph ``a $p$-value, or statistical significance, does not measure the size of an effect or the importance of a result''}  \cite{ref2}. In other words, 
statistical significance is different from
practical significance. The latter is rarely discussed in the NLP field.

\begin{figure*}
\centering
  \includegraphics[width=1\textwidth]{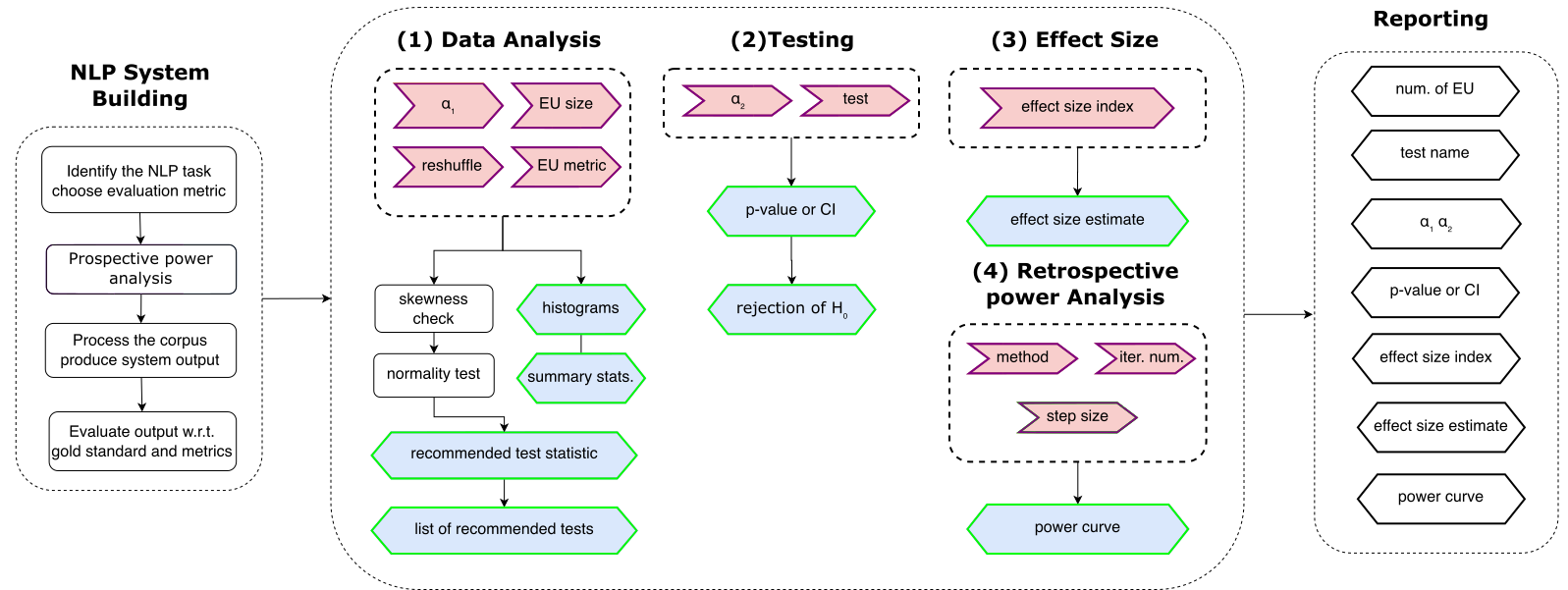}
  \caption{The three-stage procedure for comparing NLP system performance. The pink flag boxes are the parameters that users can either set or use the default values provided
  by \statsys. The blue hexagons are system output of \statsys. $\alpha_1$ and $\alpha_2$ are the significance levels for normality test and statistical significance test respectively. EU stands for evaluation unit.}
  \label{general_testing_proc}
\end{figure*}

To address this issue, we propose a three-stage procedure
for comparing NLP system performance, shown in Figure \ref{general_testing_proc}.  The first stage is building an NLP system and using \emph{prospective power analysis} to compute an appropriate sample size for test corpus. The second stage is hypothesis testing. We stress the need for data analysis to verify assumptions made by significance tests and the importance of estimating the effect size and conducting power analysis. The last stage is to report various results produced by the second stage.

To automate the process, we provide a toolkit, \statsys.
We introduce the three-stage comparison procedure (\S\ref{section2}), and then describe the
the main components (\S\ref{section3}) and implementation
details (\S\ref{section4}) of \statsys. We also present experimental results for running the system on both real-world and simulated data (\S\ref{section5}). Lastly, we compare \statsys\ with existing statistical testing toolkits (\S\ref{section6}).

\section{Comparing NLP System Performance}
\label{section2}

In this section we briefly describe the three-stage comparison procedure and define terms that are relevant to \statsys. More detail about Stage 2 can be found in \S\ref{section3}-\S\ref{section4}.

\subsection{Building an NLP System}
\label{section2-1}

The first stage is to build an NLP system, run it on test data, and compare the system output with a gold standard. The output of this stage is a list of numerical values such as accuracy or F-scores.


\begin{definition}[{\bf Evaluation unit}]\label{evalunit}
Let $(x_j, y_j)$ be a test instance. An evaluation unit (EU) $e=\{(x_j,y_j),j=1,\cdots,m\}$ is a set of test instances on which an evaluation metric can be meaningfully defined. A test set is a set of EUs.
\end{definition}

\begin{definition}[{\bf Evaluation metric}]\label{evalmetric}
Given an NLP system $A$, the evaluation metric $M$ is a function that maps an EU $e$ to a numerical value: 
\begin{equation}\label{eval-met}
  M_A(e) 
  =M\bigg(\ \{\big(\hat{y}_j,y_j\big),\ j=1,\cdots,m\}\ \bigg)
\end{equation}
where $\hat{y}_j=A(x_j)$ is the system output of $A$ given $x_j$, and $m$ is
the size of $e$ (i.e., the number of test instances in $e$).
\end{definition}

An EU may contain one or more test instances. For example, a BLEU score can be computed on one or more sentences. 
%
The EU size affects sample size, $p$-value, sample standard deviation, effect size and so on. It is therefore one of the parameters that users can set when using \statsys.


\subsection{The Comparison Stage}
\label{section2-2}

The second stage is the comparison stage 
which has four steps (see the largest box in Figure \ref{general_testing_proc}).

\subsubsection{Data Analysis}

When we compare two NLP systems $A$ and $B$, 
the output of Stage 1 is a set of pairs,  \{$(M_A(e_i), M_B(e_i))$\}, where $e_i$ is the $i^{th}$ EU, and $M_A(e)$ (similarly $M_B(e)$) is defined in Equation \ref{eval-met}.
%

Many statistical tests make certain assumptions about the sample
(e.g., normality for $t$ test), so it is important to conduct data analysis to verify those assumptions in order to choose significance tests that are appropriate for a particular sample. If the sample does not follow any known distribution, non-parametric tests should be used.

\statsys will estimate sample skewness and test for normality. Then \statsys\ will choose a test statistic (mean or median) for users and recommend a list of significance tests.

\subsubsection{Statistical Significance Testing}

The second step in Stage 2 is statistical significance
testing, using two mutually exclusive hypotheses:
the null hypothesis $H_0$ and the alternative $H_1$.
%
To compare two NLP systems, a (paired) two-sample test is usually used, though one-sample testing of pairwise difference is equivalent. \statsys\ currently only considers paired two-sample testing for numerical data. Observations within a sample are assumed to be independent and identically distributed ($i.i.d.$).

To run a significance test, users first choose the direction of the test: left-sided, right-sided or two-sided. Then, users specify the hypothesized value of test statistic difference $\delta$ and the significance level $\alpha$, which is often
set to 0.05 or 0.01 in the NLP field, and choose a test from the list. \statsys\ will calculate the $p$-value and reject $H_0$ 
if and only if $p < \alpha$.

\subsubsection{Effect Size Estimation}

In most experimental NLP papers employing significance testing, the $p$-value is the only quantity reported.
However, the $p$-value is often misused and misinterpreted.
For instance, statistical significance is easily conflated with 
practical significance; as a result, NLP researchers often run significance tests to show that the performances of 
two NLP systems are different (i.e., statistical significance),
without measuring the degree or the importance of 
such a difference (i.e., practical significance).

\newcite{ref59} noted {\it ``the null hypothesis, if taken literally, is always false in the real world."}
For instance, because evaluation metric values 
of two NLP systems on a test set are almost never exactly the same, $H_0$ that two systems perform equally is (almost) always false. When $H_0$ is false, the $p$-value will eventually approach zero in large samples \cite{ref36}. In other words, no matter how tiny the system performance difference is, there is always a large enough dataset on which the difference is statistically significant. Therefore, statistical significance is markedly different from practical significance.

One way to measure practical significance is by estimating {\em effect size}, which is defined as the degree to which the `phenomenon' is present in the population: the degree to which the null hypothesis is false \cite{ref12}.
While the need to estimate and report effect size has long been recognized in other fields \cite{ref68}, the same is not true in the NLP field.
We include several methods for estimating effect size
in \statsys\ (see \S\ref{section3-3}).
 
\subsubsection{Power Analysis}

There are two types of errors in hypothesis testing: Type I errors (false positives) and Type II errors (false negatives). 
The Type I error of a significance test, often denoted by $\alpha$, is the probability that, when $H_0$ is true, $H_0$ is rejected by the test.
 The Type II error of a significance test, usually denoted by $\beta$, is the probability that under $H_1$, $H_1$ is rejected by the test.
While Type I error can be controlled by predetermining the significance level, Type II error can be controlled or estimated by power analysis. 

\begin{definition}[{\bf Statistical power}]\label{power}
The power of a statistical significance test is the probability that under $H_1$, $H_0$ is correctly rejected by the test. The power of a test is $1-\beta$.
\end{definition}

Higher power means that statistical inferences are more correct and accurate \cite{ref32}. While power analysis is rarely used in the NLP field, it is considered good or standard practice in some other scientific fields such as psychology and clinical trials in medicine \cite{ref32}.
%
We implement two methods of conducting power analysis
in \statsys (see \S\ref{sect3-4}).

\subsection{Reporting Test Results}
\label{section2-3}

Beyond the $p$-value, it is important to report other quantities to make the studies reproducible and available for meta-analysis, including the name of significance test used, the predetermined significance level $\alpha$, effect size estimate/estimator, the sample size, and statistical power.

\section{System Design}
\label{section3}

\statsys\ is a toolkit that automates the comparison procedure.
It has four main steps, shown in the large box in Figure \ref{general_testing_proc}.
To use \statsys, users provide a data file with the NLP system performance
scores produced in Stage 1. 
\statsys\ will prompt users to either modify or use the default values
for the parameters in the pink flags and then produce the output
in the blue hexagons. 
The users can then report (some of) the output in Stage 3 of the
comparison procedure.

\subsection{Data Analysis}
\label{section3-1}

\begin{figure}
  \centering
  \includegraphics[width=1\columnwidth]{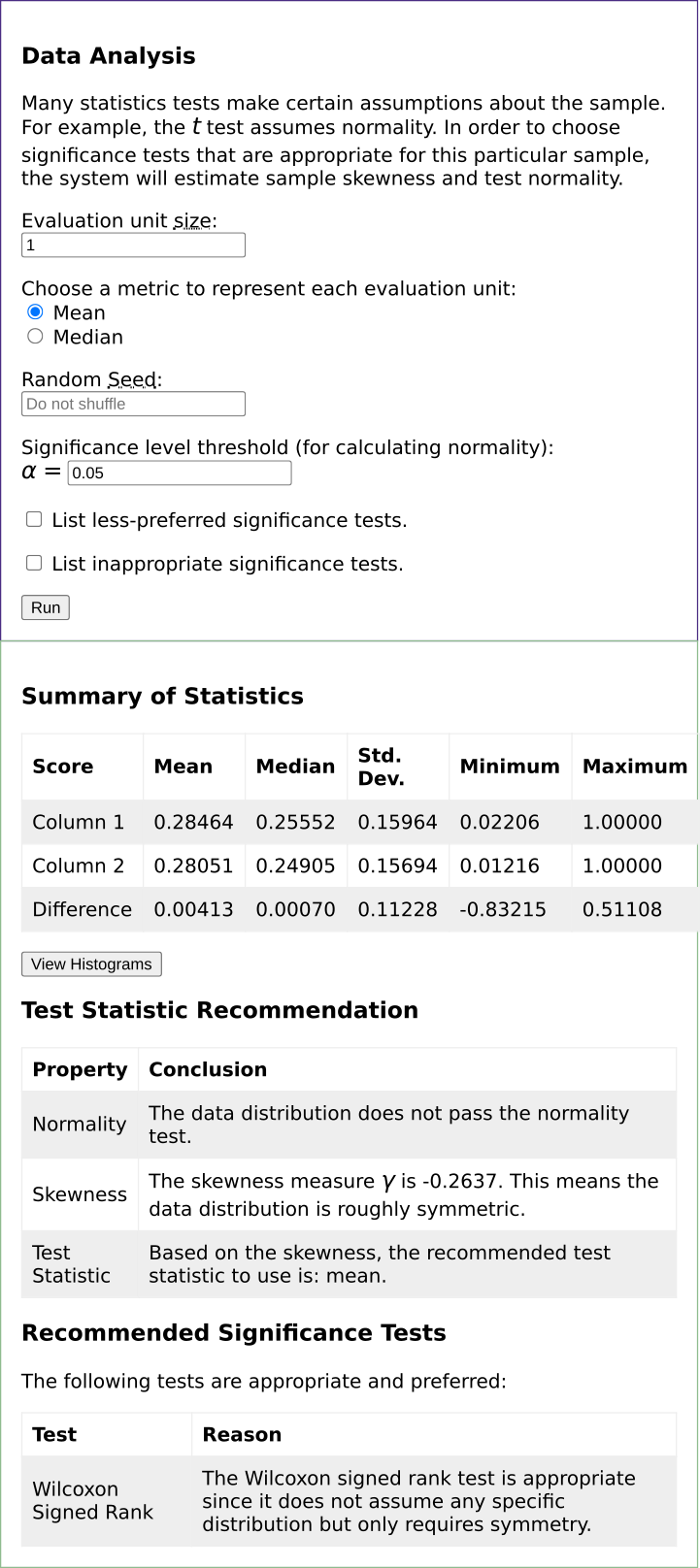}
  \caption{Screenshot of the data analysis step. The part above the {\it Run} button are
  parameters that users can set, and the part below is \statsys\ output.}
  \label{fig:data-analysis}
\end{figure}

The first step of the comparison stage is data analysis, and a screenshot
of this step is shown in Figure \ref{fig:data-analysis}. 
The top part (above the {\it Run} button in the purple box) shows the input and parameters that the user needs to provide, and the bottom part (below the {\it Run} button in the green box) shows the output of the data analysis step.

\subsubsection{The Input Data File}

To compare two NLP systems, $A$ and $B$,
users need to provide a data file where each line
is a pair of numerical values. There are two scenarios.
In the first scenario, the pair is $(u_i, v_i)$, where
$u_i = M_A(e_i)$ is the evaluation metric value
(e.g., accuracy or F-score) of an EU $e_i$ given System {\it A}
(see Equation \ref{eval-met}), and $v_i = M_B(e_i)$.

In the second scenario, if $u_i$ and $v_i$ can be calculated as 
the mean or the median of the
evaluation metric values of test
instances in $e_i$, users can
upload a data file where each line is a pair of 
($a_k$, $b_k$), where $a_k$ and $b_k$ are the evaluation metric values of 
a test instance $t_k$ given System {\it A} and {\it B}, respectively.
Users then chooses the EU size $m$  
and specifies 
whether the EU metric value should be calculated as the mean or the median of
the metric values of the instances in the EU. 
\statsys\ will use $m$ adjacent lines in the file 
to calculate $u_i$ and $v_i$.
If users prefers to randomly shuffle the lines before calculating $u_i$
and $v_i$, they can provide a seed for random shuffling.

%

\subsubsection{Histograms and Summary Statistics}

From the $(u_i, v_i)$ pairs, \statsys\ generates descriptive summary statistics (e.g., mean, median, standard deviation) and histograms of three datasets, $\{u_i\}$, $\{v_i\}$, and $\{u_i - v_i\}$, as shown in the first table
and the three histograms in Figure \ref{fig:data-analysis}.

\subsubsection{Central Tendency Measure}
Many statistical tests ($t$ test, bootstrap test based on $t$ ratios, etc) are based on the mean as the test statistic, drawing inferences on average system performance.
However, when the data distribution is not symmetric, the mean does not properly measure the central tendency. In that case, the median is a more robust measure. Another issue associated with mean is that if the distribution is heavy-tailed (e.g., the $t$ and Cauchy distributions), the sample mean oscillates dramatically. 

In order to examine the symmetry of the underlying distribution, \statsys\ checks the skewness of $\{u_i-v_i\}$ by estimating the sample skewness ($\gamma$). Based on the $\gamma$ value, we use the following rule of thumb \cite{bulmer-1979-statistics} to determine whether \statsys\ would
recommend the use of mean or median as the test statistic for 
statistical significance testing:
\begin{itemize}
  \item $|\gamma|\in[0,0.5)$: roughly symmetric (use mean)
  \item $|\gamma|\in[0.5,1)$: slightly skewed (use median)
    \item $|\gamma|\in[1,\infty)$: highly skewed (use median)
\end{itemize}

\subsubsection{Normality Test}

To choose a good significance test for \{$u_i - v_i$\}, we need to determine if the data is normally distributed.
 If it is, $t$ test is the most appropriate (and powerful) test; if not, then non-parametric tests which do not assume normality might be more appropriate.
 
If a distribution is skewed according to $\gamma$, 
there is no need to run normality test as the data is not 
normally distributed. For a non-skewed distribution,  
\statsys\ will run the Shapiro-Wilk normality test \cite{ref1}, which
is itself a test of statistical significance. The user can choose the significance 
level ($\alpha_1$ in Figure \ref{general_testing_proc}).

\subsubsection{Recommended Significance Tests}
Based on the skewness check and normality test result, \statsys\ will automatically choose a test statistic (mean or median) and recommend a list of appropriate significance tests (e.g., $t$ test if $\{u_i - v_i\}$ is normally distributed).

\subsection{Testing}
\label{section3-2}

In this step, the user 
sets the significance level ($\alpha_2$
in Figure \ref{general_testing_proc})
and chooses a significance test from the ones recommended in the previous step. If the test has any parameter (e.g., the number of trials for bootstrap testing $B$), \statsys\ will suggest a default value which can be changed by users. \statsys\ will then run the test, calculate a $p$-value (and/or provide a confidence interval), and reject $H_0$ if $p < \alpha_2$.

\subsection{Effect Size}
\label{section3-3}

Effect size can be estimated by different \emph{effect size indices}, depending on the data types (numerical or categorical) and significance tests. \newcite{ref60} defined effect size as the unstandardized difference between system performance, while \newcite{ref61} and \newcite{ref64} used the standardized difference. 




\statsys\ implements the following four indices. Once users select one or more, \statsys\ will calculate effect size accordingly and display the results.

\vspace{0.2in}
\noindent{\bf Cohen's $\bm{d}$} estimates the standardized mean difference by
\begin{equation} \label{cohend}
d=\frac{\hat{u}-\hat{v}}{\hat{\sigma}}
\end{equation}
where $\hat{v}$ and $\hat{u}$ are the sample means and $\hat{\sigma}$ denote standard deviation of $u-v$.
Cohen's $d$ assumes normality and is one of the most frequently used effect size indices. If Cohen's $d$, or any other effect size indices depending on $\hat{\sigma}$, is used to estimate effect size, the EU size  will affect the standard deviation and thus effect size estimate. 

\vspace{0.2in}
\noindent{\bf Hedges' $\bm{g}$}  adjusts the bias brought by Cohen's $d$ in small samples by the following: 
\begin{equation}\label{hedgesg}
g=d\cdot\big(1-\frac{3}{4n-9}\big)
\end{equation}
where $n$ is the size of $\{u_i-v_i\}$.

\vspace{0.2in}
\noindent{\bf Wilcoxon $\bm{r}$} is an effect size index for the Wilcoxon signed rank test, calculated as $r=\frac{Z}{\sqrt{n}}$, where
\begin{equation}
    Z=\frac{W-n(n+1)/4}{\sqrt{\frac{n(n+1)(2n+1)}{24}-\frac{\sum_{t \in T}{t^3 - t}}{48}}}
\end{equation}
Here, $W$ is the test statistic for Wilcoxon signed rank test and $T$ is the set of tied ranks.

\vspace{0.2in}
\noindent{\bf Hodges-Lehmann Estimator} \cite{ref65} is an estimator for the median. Let $w_i=u_i-v_i$. The $H\kern-0.14em L$ estimator for one-sample testing is given by
\begin{equation}
    H\kern-0.14emL=\mathrm{median}\bigg(\{(w_i+w_j)/2,\ i\ne j\}\bigg)
\end{equation}




\subsection{Power Analysis}
\label{sect3-4}

Power (Definition \ref{power}) covaries with sample size, effect size and the significance level $\alpha$. In particular, power increases with larger sample size, effect size, and $\alpha$. There are two common types of power analysis, namely {\it prospective} and {\it retrospective power analysis}, and \statsys\ implements both types.

\subsubsection{Prospective Power Analysis}
Prospective power analysis is used when planning a study (usually in clinical trials) in order to decide how many subjects are needed. In the NLP field, when one constructs or chooses a test corpus for evaluation, it will be beneficial to conduct this type of power analysis to determine how big a corpus needs to be in order to ensure that the significance test reaches the desired power level.

In \statsys, prospective power analysis is a preliminary and optional step. The user needs to provide the expected mean and standard deviation of the differences between samples, the desired power level,
and the required significance level.
\statsys\ will calculate the minimally required sample size for $t$ test via a closed form, assuming 
the normal distribution of the data. 

\begin{figure}
\centering
  \includegraphics[width=1\columnwidth]{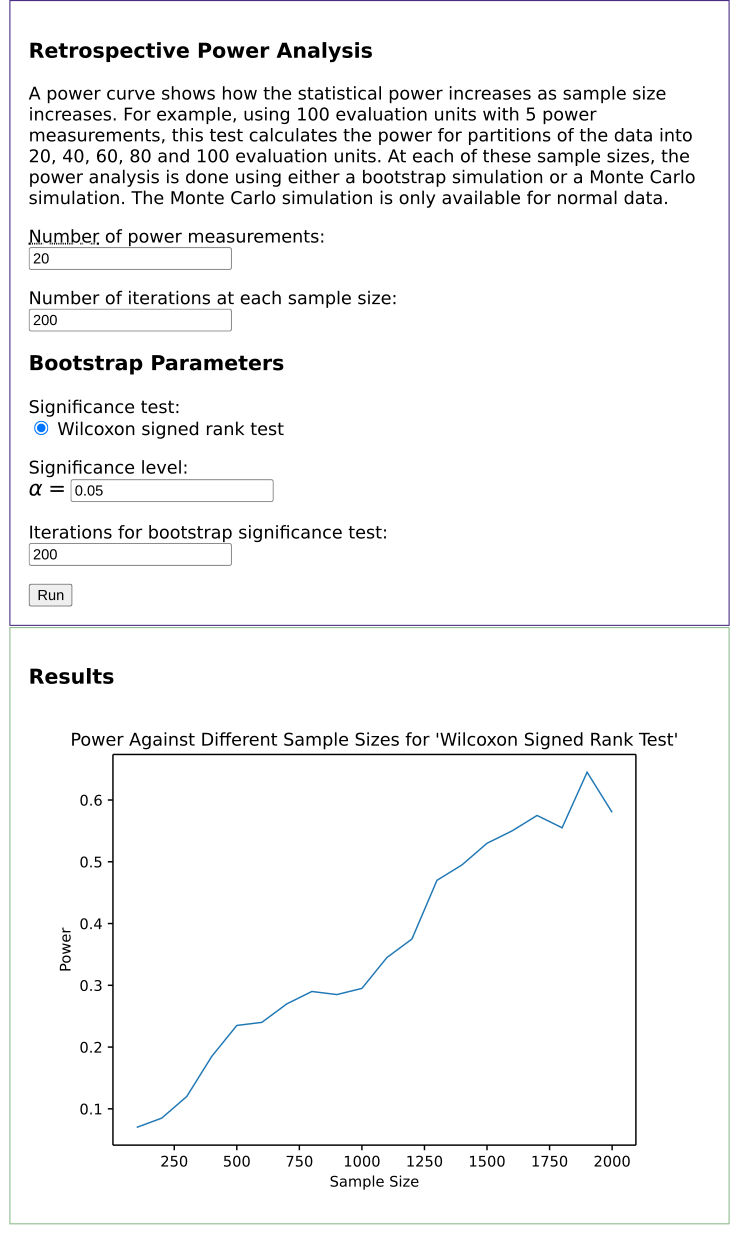}
  \caption{Screenshot for retrospective power analysis.}
  \label{fig:power-analysis}
\end{figure}

\subsubsection{Retrospective Power Analysis}
{\it Retrospective} or {\it post-hoc power analysis} is usually done after a significance test to determine the relation between sample size and power. 
There are two scenarios associated with retrospective power analysis: When the values in \{$u_i - v_i$\} are from a known distribution, one can use {\it Monte Carlo simulation} to directly simulate from this known distribution. To do this, one has to have an informed guess of the desired effect size (i.e., mean difference) via meta-analysis of previous studies. 

When the distribution of the sample is unknown \textit{a priori},
one can resample with replacement from the empirical distribution of the sample (a.k.a. the {\it bootstrap} method \cite{ref29}) to estimate the power.

\statsys\ implements both methods. Users can employ one or both;
\statsys\ will produce a figure that shows the relation between sample size and power, as in Figure \ref{fig:power-analysis}.

\section{Implementation Details}
\label{section4}


The \statsys\ graphical user interface can be run locally or on the Web. There is also a command line version. The graphical tool, the command line tool, the source code, a user manual, a tutorial video are available at  \url{nlpstats.ling.washington.edu}. We recommend using an updated Chromium-based browser.

The client-side web interface is written in HTML, CSS, and JavaScript (with JQuery). The server-side code is written in Python, using the Flask web framework. YAML is used for configuration files. KaTeX is used to render mathematical symbols. The Python code 
uses the SciPy and NumPy libraries to implement statistical tests and Matplotlib to generate the histograms and graphs.


\section{Experiment}\label{section5}

To test the output validity and speed of
\statsys, we run experiments using
both real and simulated data. 


\subsection{Real Data from WMT-2017}

The WMT-2017 shared task \cite{ref21}
reported system performance results
based on human evaluation scores;
unpaired testing (Wilcoxon rank-sum)
was used because not many sentences 
had human evaluation scores for both MT systems
that were being compared.


\begin{figure}
\centering
  \includegraphics[width=1\columnwidth]{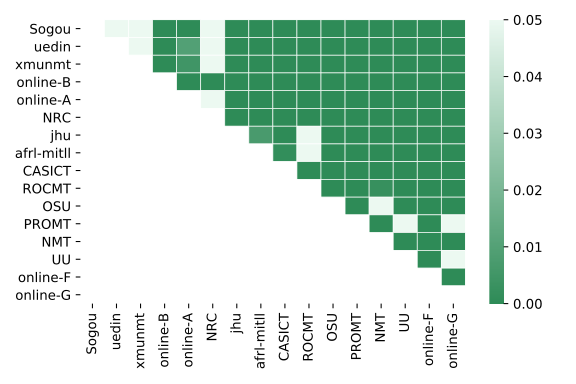}
  \caption{Heatmap of pairwise comparison for the 16 WMT-2017 Chinese-to-English MT systems. BLEU scores and Wilcoxon signed-rank test are used. $p$-values are adjusted via Bonferroni correction. Dark green cells indicate statistical significance ($p<0.05$); light green cells indicate non-significance ($p\ge0.05$).}
  \label{fig:heatmap}
\end{figure}

Because \statsys\ currently implements paired testing only, we use the Wilcoxon signed-rank test (instead of 
Wilcoxon rank-sum test) and the BLEU scores (instead of 
human evaluation scores) when comparing MT systems.
According to \newcite{ref21}, a set of 15 or more sentence-level evaluation scores constitutes a reliable measure of translation quality; thus, we set the EU size to be 15. We also reshuffled the scores before
grouping test instances into evaluation units.

Figure \ref{fig:heatmap} shows the results of pairwise comparisons among all 16 Chinese-to-English MT systems (120 system pairs in total). The heatmap is similar 
to the comparison results in \newcite{ref21}  (see Figure 5 in that paper). The minor differences of the two heatmaps
are due to different evaluation metrics (BLEU vs. human scores), the significant tests (Wilcoxon signed-rank vs. Wilcoxon rank-sum), and the numbers of EUs (more test sentences have BLEU scores than human evaluation scores).

\begin{figure}
\centering
  \includegraphics[width=1\columnwidth]{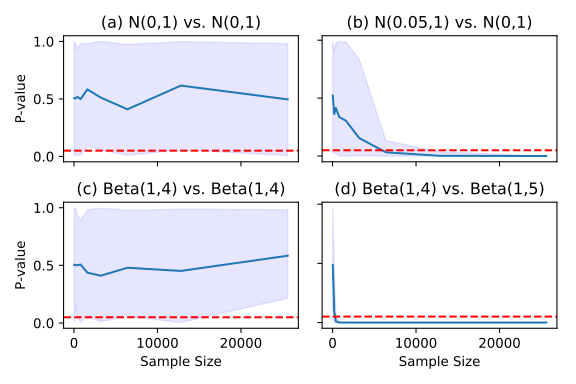}
  \caption{Plots of $p$-value against sample size. Figure (a) and (b) use two samples with normal distribution, while (c) and (d) use Beta distribution. $H_0$ should be true for (a) and (c) and false for (b) and (d). We run $t$ test for (a) and (b), and Wilcoxon signed-rank test for (c) and (d). The red dotted line stands for the threshold $\alpha=0.05$. The light purple shade depicts the range of $p$-values. The solid blue line denotes the mean of $p$-values for each sample size.}
  \label{fig:sim-exp-norm}
\end{figure}

\subsection{Simulated Data}
We also run simulation experiments on \statsys\ to validate the testing results. Here, we conduct two-sided, paired testing, varying sample size from 30 to 25,000, each with 20 iterations of tests to obtain a range of $p$-values. 
As shown in Figure \ref{fig:sim-exp-norm}, when $H_0$ is true (see Fig \ref{fig:sim-exp-norm}(a) and \ref{fig:sim-exp-norm}(c)), $p$-values range freely in $(0,1)$. When $H_0$ is false (see \ref{fig:sim-exp-norm}(b)
and \ref{fig:sim-exp-norm}(d)),
$p$-values approach zero as sample size increases, as expected. The fast convergence to zero in \ref{fig:sim-exp-norm}(d) may be due to the small variance of the differences between the two Beta samples ($\approx 0.046$), even though the difference between sample medians is small ($\approx 0.02$).
In contrast, \ref{fig:sim-exp-norm}(b) converges to zero much more slowly due to the large variance.


\section{Related Work}\label{section6}
\newcite{ref3} made an accompanying package available \footnote{\url{https://github.com/rtmdrr/testSignificanceNLP}} for hypothesis testing. This package includes functionalities such as testing for normality, $t$ testing, permutation/bootstrap testing, and using McNemar's test for categorical data. \statsys\ implements all the aforementioned tests except McNemar's test. In addition, \statsys\ offers data analysis, effect size estimation, power analysis and graphical interface. 

\statsys\ is based on the frequentist approach to hypothesis testing. \newcite{Sadeqi-2020-hypobayes} developed a Bayesian system\footnote{\url{https://github.com/allenai/HyBayes}} which uses the Bayes factor to determine the posterior probability of $H_0$ being true or false.

\section{Conclusion}

While statistical significance testing has been commonly used to compare NLP system performance,
a small $p$-value alone is not sufficient
because statistical significance is different from practical significance.
To measure practical significance, we recommend estimating and reporting of effect size. It is also necessary to conduct power analysis to ensure that the test corpus is large enough to achieve a desirable power level.
%
We propose a three-stage procedure for comparing NLP system performance,
and provide a toolkit, \statsys, to automate the testing stage of the procedure.
For future work, we will extend this work to hypothesis testing with 
multiple datasets or multiple metrics. 



\bibliographystyle{acl_natbib}
\bibliography{biblio}

\end{document}